\documentclass[11pt]{article}

\usepackage[preprint]{acl}

\usepackage{times}
\usepackage{latexsym}

\usepackage[T1]{fontenc}

\usepackage[utf8]{inputenc}

\usepackage{microtype}

\usepackage{inconsolata}

\usepackage{graphicx}

\usepackage[utf8]{inputenc} 
\usepackage[T1]{fontenc}    
\usepackage{hyperref}       
\usepackage{url}            
\usepackage{booktabs}       
\usepackage{amsfonts}       
\usepackage{nicefrac}       
\usepackage{microtype}      
\usepackage{xcolor}         
\usepackage{graphicx}
\usepackage{times}
\usepackage{latexsym}
\usepackage{pdfpages}
\usepackage{tabularx}
\usepackage{array}
\usepackage{multirow}
\usepackage{ulem}
\usepackage{pifont}
\usepackage{longtable}
\usepackage{multirow}

\usepackage{amsmath}
\newcommand{\xmark}{\text{\ding{55}}}
\usepackage{adjustbox}
\usepackage{wrapfig}
\usepackage{multirow}
\usepackage{makecell}

\definecolor{darkblue}{RGB}{46,25, 110}

\title{Outside Knowledge Conversational Video (OKCV) Dataset - Dialoguing over Videos}

\author{
  \textbf{Benjamin Reichman}$^{1}$ \quad \textbf{Constantin Patsch}$^{2}$ \quad \textbf{Jack Truxal}$^{1}$ \quad \textbf{Atishay Jain}$^{1}$ \quad \textbf{Larry Heck}$^{1}$ \\
  $^1$Georgia Institute of Technology \quad $^2$Technical University of Munich \\
  \texttt{\{bzr,jtruxal6,atishay.jain,larryheck\}@gatech.edu}\\
  \texttt{constantin.patsch@tum.de}\\
}

\begin{document}
\maketitle
\begin{abstract}
In outside knowledge visual question answering (OK-VQA), the model must identify relevant visual information within an image and incorporate external knowledge to accurately respond to a question. Extending this task to a visually grounded dialogue setting based on videos, a conversational model must both recognize pertinent visual details over time and answer questions where the required information is not necessarily present in the visual information. Moreover, the context of the overall conversation must be considered for the subsequent dialogue. To explore this task, we introduce a dataset comprised of $2,017$ videos with $5,986$ human-annotated dialogues consisting of $40,954$ interleaved dialogue turns. While the dialogue context is visually grounded in specific video segments, the questions further require external knowledge that is not visually present. Thus, the model not only has to identify relevant video parts but also leverage external knowledge to converse within the dialogue. We further provide several baselines evaluated on our dataset and show future challenges associated with this task. The dataset is made publicly available here: \href{https://github.com/c-patsch/OKCV}{https://github.com/c-patsch/OKCV}
\end{abstract}

\section{Introduction}
Much of the human experience is mediated by continuous visual perception. Consequently, understanding videos—analogous to this continuous perception—is a crucial task for artificial intelligence. Intelligent comprehension of the world includes the ability to filter relevant aspects of visual information, engage in dialogue about it, and link it to knowledge that extends beyond immediate perceptual content. These are skills that are beyond current large vision foundation models.

In Visual Question Answering (VQA), models are tasked with answering questions about the content of an image. Outside knowledge-based VQA not only relates the question to visual cues but rather also extends it to knowledge that is not present within the image~\cite{marino2019ok,reichman2023outside}. Thus, current approaches utilize external knowledge bases like Wikidata or Wikipedia~\cite{wang2015explicit, gao2022transform, kat, lin2022revive} to answer questions that require information that goes beyond the visual content of the image. 
Due to the promising reasoning capabilities of LLMs, recent approaches leverage models like GPT-3 and GPT-4~\cite{brown2020language,gpt4} to further improve answers with respect to the knowledge inherently present in the pretraining datasets. These approaches often ~\cite{shao2023prompting, yang2022empirical} use image captioning models to transfer the visual information present in the image to the text domain.
A further extension of the VQA is the Video QA task. As humans generally perceive temporally evolving visual information in the form of videos, Video QA~\cite{fu2021violet, zellers2021merlot} provides an analog to this by visually grounding questions in terms of video data. VQA dialogue systems depending on such visual grounding, are then capable of having a conversation about the image or video content~\cite{nie2019multimodal, engin2021hidden}.

\begin{figure*}[t!]
\centering
\includegraphics[width=\textwidth]{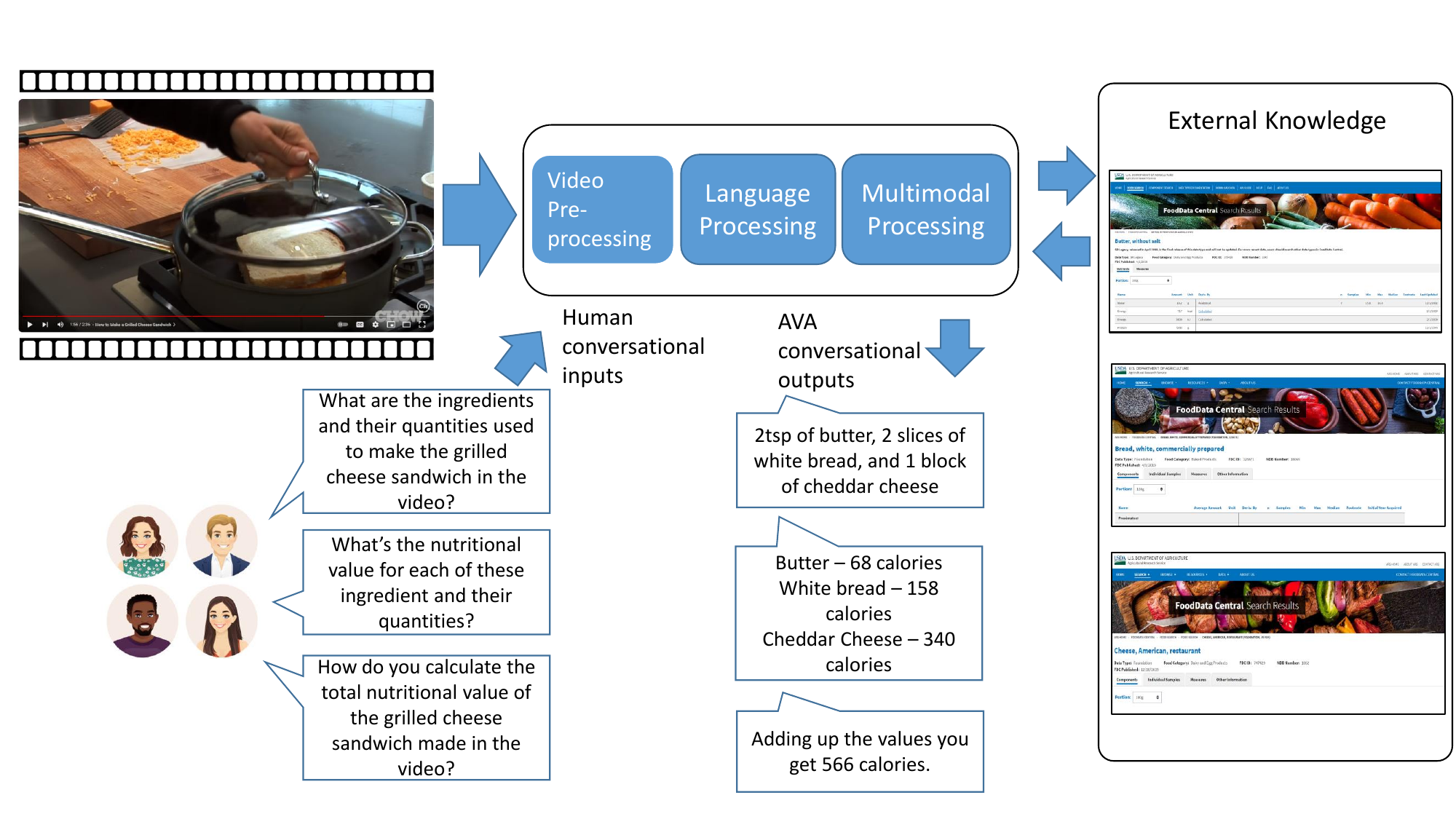}
\caption{An example conversation from the OKCV dataset.}
\label{fig:example_graph}
\end{figure*}

Our dataset utilizes videos for visual grounding, requiring models to process information both spatially and temporally. This necessitates that models discern which parts of a video are pertinent to answering a question.
With each video, we provide corresponding dialogues requiring the model to process the visual information present in the video, retrieve and reason over external knowledge, and be able to interact with users within a dialogue setting.
Thus, compared to previous VQA and outside knowledge VQA datasets, our dataset tests a model's capability to understand visual scenes and reason about it with additional information that is not present in the visual information. 
Providing dialogues instead of one-off QA like previous outside knowledge VQA datasets enables testing a model's ability for extended reasoning capabilities over a video and the ability to mimic natural conversational scenarios. Thus, a model has to consider visual cues, external knowledge not present in the video, and the overall context of dialogue.

To address these challenges, we introduce a new dataset, OKCV, which consists of $2,017$ videos for visual grounding, accompanied by $5,986$ human annotated dialogues consisting of $40,954$ interleaved dialogue turns. These conversations evolve around the visual context but require external knowledge to effectively complete a dialogue turn. Consequently, our dataset challenges the capabilities of video-language models in terms of their spatiotemporal, factual, and commonsense reasoning as well as their dialogue management skills. Thus, resulting dialogue systems should be able to have natural comprehensive conversations about videos extending beyond their internal knowledge bases.

\begin{table*}[t]
\centering
\small
\adjustbox{max width = \textwidth}{
\scalebox{0.97}{\begin{tabular}{lcccccccc}
\toprule
\multirowcell{2}{\textbf{Dataset}} & \multirowcell{2}{\textbf{VQA-Type}} & \multirowcell{2}{\textbf{\# Dial.}} & \multirowcell{2}{\textbf{Vis.}} & \multirowcell{2}{\textbf{Text}} & \multirowcell{2}{\textbf{Temp.}} & \multirowcell{2}{\textbf{Know.}} & \multirowcell{2}{\textbf{Conv.}} & \multirowcell{2}{\textbf{Human} \\ \textbf{Annot.}}\\[9pt]
\midrule
OK-VQA~\cite{marino2019ok}& ImageKB & - & \checkmark &  \checkmark & \xmark & \checkmark & \xmark & \checkmark\\

Visual Dialogue~\cite{das2017visual} & Image-Conv & 140k & \checkmark &  \checkmark & \xmark & \xmark & \checkmark & \checkmark\\

TVQA~\cite{lei2018tvqa} & Video & - & \checkmark &  \checkmark & \checkmark & \xmark & \xmark & \checkmark\\

TVQA+~\cite{lei2019tvqa+} & Video & - & \checkmark &  \checkmark & \checkmark & \xmark & \xmark & \checkmark\\

AVSD~\cite{alamri2019audio} & Video-Conv & 11.8k & \checkmark &  \checkmark & \checkmark & \xmark & \checkmark & \checkmark\\

KnowIT-VQA~\cite{garcia2020knowit}& VideoKB & - & \checkmark &  \checkmark & \checkmark & \checkmark & \xmark & \checkmark\\

VideoChat~\cite{li2023videochat} & Video-Conv & 4k & \checkmark &  \checkmark & \checkmark & \xmark & \checkmark & \xmark\\

OKCV (ours) & VideoKB-Conv & 5.7k & \checkmark &  \checkmark & \checkmark & \checkmark & \checkmark & \checkmark\\


\bottomrule
\end{tabular}
}
}
\caption{Comparision of the dataset characteristics number of dialogues (\# Dial.), visual grounding (Vis.), text (Text), temporal information in the form of videos (Temp.), required external knowledge (Know.), conversational structure (Conv.) and human-annotated data (Human Annot.). VQA type indicates the type of visual grounding, required external knowledge and dialogue data structure.}
 \label{tab:datasets}
\end{table*}

\section{Related Work}
\subsection{Visual Question Answering (VQA)}
In VQA, models must align text and vision representations to reason about the visual content of an image. VQA datasets generally contain question-answer pairs that are assigned to specific images~\cite{antol2015vqa, gao2015you, goyal2017making}. In this task, the model has to reason about the visual content of the image and be able to answer related questions. 
Knowledge-based VQA expands on this by requiring models to consider facts beyond the visual content of an image~\cite{wang2017fvqa, shah2019kvqa, marino2019ok, reichman2023outside}.
VideoQA datasets take this further by grounding the model in video content rather than static images ~\cite{tapaswi2016movieqa, lei2019tvqa+, garcia2020knowit}. Thus, tasking the model with extracting not only spatially relevant information from frames but also temporally important information throughout the video duration.

There are several existing datasets in the VQA and Video QA area that require external knowledge to answer the questions contained in them. In the image modality, datasets such as OK-VQAv2, A-OKVQA, and KVQA feature a wide array of questions that necessitate this additional knowledge \cite{reichman2023outside,marino2019ok,aokvqa,kvqa}. In the video modality, there is the KnowIT VQA and KnowIT-X VQA datasets~\cite{garcia2020knowit,knowitxvqa} introduce human-generated question-answer pairs, that integrate visual, textual, and temporal reasoning with knowledge-based questions, where the required knowledge-base is restricted to the sitcom. However, these datasets are restricted to questions about popular sitcoms, The Big Bang Theory and Friends, respectively, limiting the applicability of the dataset. NEWSKVQA \cite{NEWSKVQA}, another dataset in this domain, is restricted to questions about the news and employs entirely automated methods for question-answer generation. Beyond single-turn question-answering datasets, there is the recent creation of the VideoChat dataset \cite{li2023videochat} which starts to fill the gap in the dialogue space. However, this dataset contains mostly visual-based dialogues without the need for outside knowledge and relies entirely on machine-generated content. 

Compared to other external knowledge-based VQA and VideoQA datasets, our dataset is a human-generated dataset that focuses on the pairing of video understanding, external knowledge retrieval and integration, and conversational reasoning. Furthermore, it is not confined to any specific area of focus, offering a broader application potential across various domains.
One essential key difference from other datasets is that we decompose one high-level question into several dependent subquestions, which ensures the semantic consistency of the dialogue. This is further explained in Section \ref{subsec:gpt}.

\subsection{Dialogue Datasets}
Open-domain dialogue datasets typically consist of textual conversations, where a model is required to consider the dialogue context that evolves over the conversation~\cite{lison2016opensubtitles2016, sordoni2015neural, multiwoz, byrne2019taskmaster}. The Taskmaster-1 dataset~\cite{byrne2019taskmaster}, features goal-oriented dialogue without scripts or restricted knowledge bases. 
The CommonsenseQA dataset~\cite{talmor2019commonsenseqa} involves commonsense question answering, sourced from Conceptnet~\cite{speer2017conceptnet}, featuring multiple-choice questions crafted by crowd-workers.

Beyond purely text-based datasets, visually grounded conversational datasets enable models to engage in discussions about the content of images or videos~\cite{das2017visual, meng2020openvidial}. The Visual Dialogue dataset~\cite{das2017visual} pairs images with dialog histories modeled after two-person chats, requiring models to answer questions related to the visual content within the context of a conversation. Similarly, the OpenViDial dataset~\cite{meng2020openvidial} pairs dialogue turns with visual contexts from movies and TV series, necessitating both visual and textual conversational reasoning. The Audio Visual Scene-aware Dialog (AVSD) dataset~\cite{alamri2019audio} consists of dialogs and video summaries designed to enable models to generate natural responses using video, audio, and dialog history. In contrast, to previous visually grounded dialogue datasets, our dataset focuses on spatial and temporal visual grounding and includes human-annotated conversations that relate to the video while also requiring external knowledge to keep the conversation.


\section{Dataset Collection}
\label{sec:dataset_collection}
We present the Outside-Knowledge Conversational Videos (OKCV) Dataset, a dataset of video-based dialogues that require outside knowledge for correct completion. This dataset extends the open-knowledge VQA task by incorporating a temporal dimension to both the input and output modalities. Rather than using static images, our dataset requires vision models to process videos. Instead of a single question-answer turn, the system must handle evolving dialogue states while also retrieving and integrating outside information. 

Collecting human-annotated data for either video or dialogue tasks is costly, time-consuming, and labor-intensive. To reduce the burden on human annotators and reduce cost-whether monetary, time, or labor-related—we leveraged the existence of pre-existing video datasets and different LLM capabilities. This section details the pipeline used for collecting and annotating the dataset.

\subsection{Video Dataset Selection}
The first step in creating the OKCV Dataset involved collecting videos around which dialogues would be constructed. The objective of the video selection phase was to assemble a diverse set of videos conducive to the creation of interesting and varied dialogues. An additional requirement was that the chosen videos must come with textual descriptions and transcripts. The availability of transcripts rather than just captions, was essential; transcripts provide more detailed information about the video content,  which is crucial for the subsequent stages of the dataset creation pipeline.

These criteria led us to select the QuerYD dataset\footnote{From email discussions with the first author of the QuerYD publication, the videos are from YouTube which are covered by fair use. The QuerYD dataset is released for research purposes, which this work is.} as the base for our dataset \cite{oncescu2021queryd}. The videos Queryd vary in length, encompass a breadth of topics and employ an extensive vocabulary. Crucially, transcripts were available for these videos. Out of the 2593 videos in Queryd we used $2,017$ videos. We opted to only include videos ranging from 30 seconds to 10 minutes in length, striking a balance between dialogue and the costs associated with collecting data over long videos.

\subsection{GPT Question Generation}\label{subsec:gpt}
With the videos selected, the next step was to generate the dialogues. Relying solely on human annotators to watch the videos and create dialogues would be extremely labor-intensive and costly. Consequently, the availability of video transcripts was crucial. Instead of humans crafting the entire dialogue from scratch, we employed a combination of the transcripts and a LLM (GPT-4) to create partial dialogue drafts. This approach reduced the workload for human annotators in creating the dataset.

The partial dialogue drafts created by GPT-4 only contained the questioner's side of the conversation. This was done to preserve human reasoning and judgment in the dataset creation. Creating questions is an open-ended, challenging, and time-consuming task for humans. Moreover, human-generated questions may not consistently meet the specific criteria set for our dataset. By leveraging the video transcription and GPT-4’s internal knowledge, we were able to generate reasonable questions for humans to answer that adhered to certain prescribed parameters:
\begin{enumerate}
    \item The question must be related to the video.
    \item The question must require outside knowledge.
    \item The question should require multiple frames to answer.
\end{enumerate}

To prompt the LLM to create partial dialogues, an overarching strategy for dialogue construction was necessary. We experimented with two different dialogue strategies: bottom-up and top-down. The bottom-up strategy involved generating many individual, simpler questions, from which the top 5-6 were selected and stitched together into a dialogue. In contrast, the \textbf{top-down dialogue strategy} started with one complex question, which was then decomposed into smaller, simpler questions to form a dialogue. After conducting several qualitative experiments with both approaches, we opted for the \textbf{top-down strategy}. It was observed to yield more coherent dialogues that more frequently required the integration of outside knowledge.

Through qualitative experimentation, we refined our prompting strategies for GPT-4. Initially, we prompted GPT-4 in a zero-shot manner, providing only instructions on how to generate top-down questions. This method resulted in high variability in question quality. In the next iteration, we incorporated high-quality questions from the previous trials into the prompt for GPT-4, adopting a few-shot prompting approach that stabilized question quality. Further improvements were achieved when GPT-4 was tasked with simultaneously creating and decomposing the questions into smaller subquestions. This encouraged the model to generate more structurally decomposable questions. The final quality enhancement involved prompting GPT-4 to explicitly explain how outside knowledge and multiple frames were necessary to answer the questions it generated. For details on the specific prompts used to generate the final dataset, please refer to the supplementary material.

By generating one high-level multifaceted question and decomposing it into several subquestions that subsequently depend on answering the respective previous one, we ensure semantic consistency of the overall dialogue structure. This is a key difference from other datasets that only rely on independent question-answer pairs.

With the established prompting strategy, GPT-4 was tasked with creating ten questions for each video. From these, three questions were randomly selected for human annotation.

\subsection{Human Annotation}

\begin{table}[pt]
{
\centering

\scalebox{0.85}{\begin{tabularx}{{0.55\textwidth}}{ll}
\toprule
\multicolumn{2}{c}{\textbf{OKCV Characteristics}} \\
\midrule
\# of Videos & $2,017$  \\
\# of Dialogues & $5,986$ \\
\# of Dialogue Turns & $40,954$  \\
Average \# of Dialogue Turns/Video & $6.8$\\
\# of Dialogues with Sources & $550$ \\
\# of Sources & $641$ \\
\# of Valid Temporal Certificates Collected & $4,525$ \\
Average Temporal Certificates Length & $42.3$s \\
Average Video Length & $178$s \\
\bottomrule
\multicolumn{2}{c}{\centering\textbf{Linguistic Characteristics}} \\
\midrule
\# of Tokens & $450,260$  \\
Total Vocab Size & $17,462$ \\
\# Nouns & $9,789$  \\
\# Verbs  & $4,917$ \\
\# Adjectives & $3,012$ \\
\# Adverbs & $732$ \\
Sentence Length/Turn & $10.5$ \\
\bottomrule
\end{tabularx}
}
\caption{Quantitative insight into the OKCV dataset.}
\label{tab:knowledge_added}}
\end{table}


With the videos selected and partial dialogues drafted by GPT-4, the final step involved completing the dialogues. This phase required human involvement. GPT-4 served as a preliminary source of dialogue content, generating responses based on its interpretation of the video transcripts. Human annotators from English-speaking countries, capable of viewing and processing the videos, then refined and expanded these dialogues, ensuring they were coherent and contextually appropriate.

We provided the turkers with an interface (see supplementary material) to view and complete the dialogues. The interface was designed to simulate a phone conversation, enhancing the intuitiveness of the task. In this interface, turkers would watch the video and then complete the three partial dialogues generated by GPT-4 for that specific video. The user had the option to modify GPT-4’s initial generation if they believed changes would improve the conversational flow or be more faithful to the video. The turkers were required to answer the questions presented in the partial dialogues, effectively fulfilling their role in the simulated conversation.

In addition to completing the dialogues, turkers were requested to provide two additional pieces of information for each dialogue. 
First, turkers were instructed to list any sources they used or deemed relevant for answering the dialogue questions. The aim was to establish a ground truth for any retrieval system trained or tested on our dataset. Secondly, they were required to create temporal certificates for the OKCV dataset, a certification method described by the EgoSchema dataset \cite{egoschema}. This allows us to characterize how much of the video content is necessary to adequately respond to the dialogues generated.

\section{Dataset Analysis}
To gain a deeper insight into the characteristics of the OKCV dataset, this section offers a detailed analysis focusing on linguistic complexity and the extent of video coverage required to adequately respond to the dialogues.

The OKCV dataset consists of a total of $2,017$ videos, with each video featuring approximately 2.96 dialogues, amounting to a total of $5,986$ dialogues. Each dialogue consists of an average of 6.8 turns, resulting in 40,954 dialogue turns across the dataset. In order to investigate how many questions rely on external knowledge compared to only vision-based information, we used the DeepSeek-R1 model~\cite{guo2025deepseek} distilled into a Llama model with 8 billion parameters. The results showed that $57.93\%$ of the questions require external knowledge and cannot be solely solved with visual information.
Additionally, by manually checking a randomly sampled subset of the data we verified the dialogue nature and semantic dependencies between multiple turns.

The videos, as discussed in Section \ref{sec:dataset_collection}, originate from the Queryd dataset, which itself sources the videos from YouTube \cite{oncescu2021queryd}. The Queryd dataset, and thus the OKCV dataset, cover a wide range of topics \cite{oncescu2021queryd}. This diversity is reflected in the linguistic characteristics of the dataset. Using the Spacy library for tokenization and part-of-speech analysis \cite{spacy2}, we found that the OKCV dataset possesses a total vocabulary size of $17,462$, which includes $9,789$ nouns and $4,917$ verbs. This extensive vocabulary range provides a rich array of concepts for models trained on this dataset to learn and visually interpret.

\begin{figure*}[t!]
\centering
\includegraphics[width=0.9\textwidth]{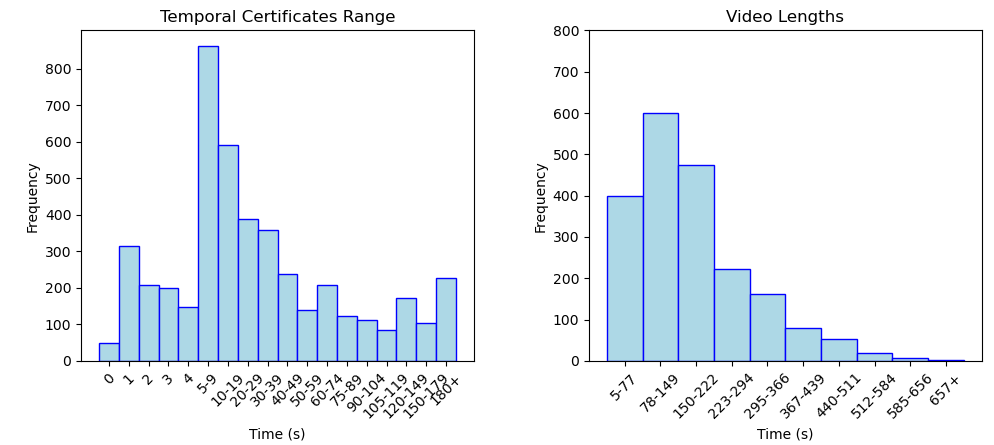}
\caption{Distribution over the temporal certificate ranges as well as the video lengths in OKCV.}
\label{fig:time_stamps}
\end{figure*}

\begin{figure*}[t!]
\centering
\includegraphics[width=0.85\textwidth]{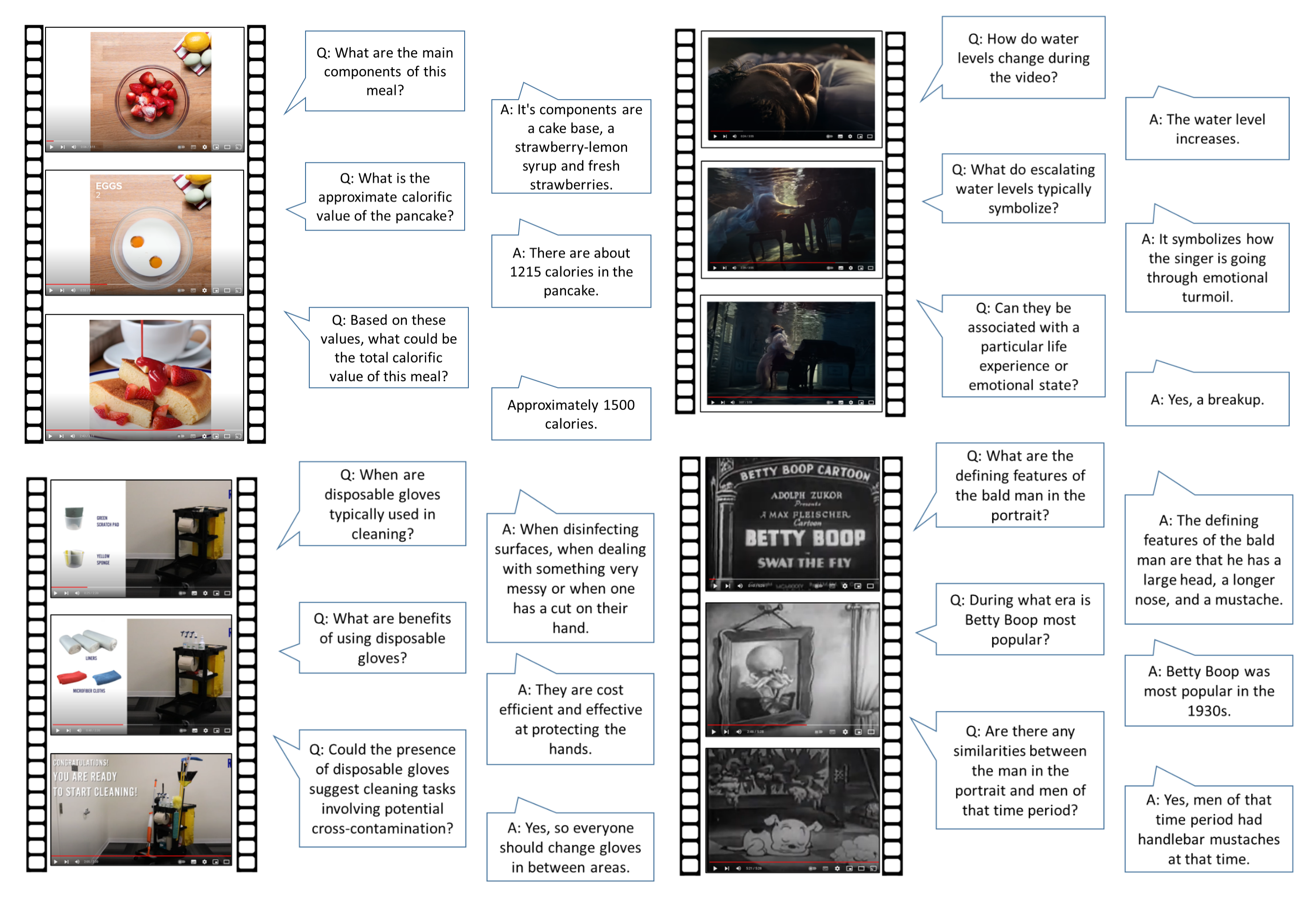}
\caption{Examples of visually grounded conversations and corresponding video snippets from the OKCV dataset.}
\label{fig:qual_examples_}
\end{figure*}

Figure \ref{fig:time_stamps} illustrates the distribution of temporal certificates and video lengths within the dataset. The videos vary widely in duration, with an average length of $178$ seconds. Temporal certificates, indicating the portion of the video needed to answer the dialogues, are primarily clustered between $5$ and $30$ seconds. However, a significant number of videos require longer viewing, with an average temporal certificate of $42$ seconds. Although, on average, only about a quarter of each video is necessary to complete the dialogue, this data suggests that for the OKCV dataset, the ability to process and comprehend longer video sequences is essential for successfully completing the tasks. In Figure \ref{fig:qual_examples_}, we show example dialogue-style question-answer pairs of our dataset where subsequent questions relate to previous answers.



\section{Baseline and Results}


In this section, we explore various baselines for the OKCV dataset. We evaluate four base model architectures, applying zero-shot prompting to all of them. 
We investigate 
three different combinations of external knowledge. Paired with the baselines we also explore three different evaluation strategies to assess model performance. 

We experimented with several backbone models to serve as the drivers for our dialogue system: Llama2-7B, Llama3-8B, Mistral-7B, Phi-3, and Video-Llama. These LLMs are highly proficient in general text-based dialogue and possess extensive stored knowledge, which can be leveraged in dialogue contexts. Llama2-7B was selected as it has a robust open-source community, which has developed various augmentations. As we were establishing baselines with Llama2, Llama3 was released, prompting us to include Llama3-8B in our baselines for comparative analysis. Phi-3 and Mistral-7B were selected as they both also represent high-quality open-source models. Video-Llava was selected as it is a highly-performant video language model. As this dataset engages in both the vision and language modalities, it was necessary to test a model that is able to engage with both aspects of the dataset natively. 

For the second knowledge setting, we took inspiration from the PICA method used in the OK-VQA task \cite{yang2022empirical}. This method separates the vision component of the task from the natural language processing element of the task. In the PICA method, the images are converted into language descriptions using a captioning model, and these captions, along with the question, are then fed into an LLM. The PICA method did this exclusively in a zero-shot setting. In our experiments, we reproduce this approach using a vision-language model called mPLUG-2~\cite{xu2023mplug}. The dialogue model utilizes the video caption and the current dialogue state to predict the subsequent dialogue stage. We experimented with both prompting the model in a zero-shot setting and fine-tuning the models to complete the dialogue using a caption derived from the video.

\begin{table*}[t]
\centering
\captionsetup{justification=centering}
{
\small
\begin{tabular}{c|c|c|c|c}
\toprule
 Backbone Model & Knowledge & Bleurt & BartScore & Prometheus-2  \\
\midrule
LLama2-7B & Nothing & $0.49$ & $0.034$ & $3.58$ \\
\hline
LLama2-7B & Caption & $0.47$ & $0.033$  & $3.37$ \\
\hline
LLama2-7B & Caption+Knowledge & \textbf{$0.58$} & \textbf{$0.117$} & $2.27$ \\
\hline
LLama3-8B & Nothing & $0.52$  & $0.035$ & $3.10$ \\
\hline 
LLama3-8B & Caption & $0.50$ & $0.034$ & $2.95$ \\
\hline 
LLama3-8B & Caption+Knowledge & $0.49$ & $0.036$ & $2.03$ \\
\hline 
Mistral-7B-Instruct-v0.3 & Nothing & $0.52$ & $0.041$ & $3.66$ \\
\hline
Mistral-7B-Instruct-v0.3 & Caption & $0.52$ & $0.040$ & $3.53$ \\
\hline
Mistral-7B-Instruct-v0.3 & Caption+Knowledge & $0.51$ & $0.040$ & $3.29$ \\
\hline 
Phi-3-Small & Nothing & $0.49$ & $0.039$ & \textbf{$3.74$} \\
\hline
Phi-3-Small & Caption & $0.49$ & $0.039$ & $3.03$ \\
\hline
Phi-3-Small & Caption+Knowledge & $0.49$ & $0.040$ & $3.53$ \\
\hline 
Video-Llava & Nothing & $0.59$ & $0.047$ & $3.30$ \\
\hline
Video-Llava & Knowledge & $0.58$ & $0.045$ & $3.41$ \\
\hline \hline
Gemini-1.5-pro* & Nothing & $0.98$ & $0.016$ & $2.39$ \\
\bottomrule
\end{tabular}
}
\caption{Zero-shot results on the OKCV dataset.\\ *Due to the cost of running Gemini only $45\%$ of videos were evaluated.}
\label{tab:baseline_results_zero_shot}
\end{table*}

The final knowledge setting we employed was also inspired by approaches used in the OK-VQA task, specifically the KAT and Revive methods \cite{kat,lin2022revive}. In this method, a dual-pronged approach was used. The first prong uses a multi-modal joint embedding space to retrieve textual knowledge entities, effectively bridging information found in the video into natural language while providing extra context and outside knowledge. This retrieval was facilitated by the Clip4Clip model \cite{clip4clip}. Due to the limited input sequence length of CLIP, we used Wikidata as our retrieval corpus \cite{wikidata}, specifically leveraging the subset previously utilized by related works in the literature \cite{kat,lin2022revive,crossmodalretrieval}. The knowledge retrieved by the Clip4Clip model along with the captions generated earlier, were inputted into our LLM to help guide its generation of the next dialogue state. Unlike the original KAT and Revive methods, we did not include an implicit knowledge module. The LLMs being used to drive the dialogue are comparable to the implicit knowledge modules those methods used. The LLMs chosen utilized a blend of their implicit knowledge and the prompted knowledge to generate the next dialogue state. This combines the functions of the implicit knowledge module and knowledge aggregators in the KAT and Revive frameworks and thus simplifies the overall framework. We experimented with this knowledge setting in both the zero-shot and fine-tuned settings.

Evaluating dialogue models presents significant challenges due to the multitude of valid potential responses in a dialogue. Even with a "ground truth" reference statement, there are multiple expressions of the same idea that convey similar meanings. Given that 
we are assessing 
zero-shot completions, using automated metrics like ROUGE or BLEU may not be appropriate, as they are not designed to evaluate semantic equivalence. Models prompted in a zero-shot manner are not designed to produce responses in the same style as our dataset though they may still generate semantically equivalent outputs. We therefore opted for trained automated evaluation metrics that are better suited to capture semantic accuracy. 

To capture a variety of perspectives we selected three different scoring systems: Bleurt, BartScore, and Prometheus-2 \cite{bleurt1, bleurt2,bartscore,prometheus2}. All three are reference-based metrics. Bleurt is a supervised regression-based method trained to mimic human judgment by judging the distance between a candidate text and a reference text. BartScore, an unsupervised metric, evaluates the likelihood of the candidate text given the hypothesis text and vice versa. Lastly, Prometheus-2 is a supervised evaluator model that provides verbal feedback and scores outputs on a 1-5 scale using the Likert system, according to a user-provided rubric. The rubric we used in the evaluation is detailed in Appendix \ref{app:promethrubric}. By using multiple trained metrics for model assessment, we hope to capture a more nuanced understanding of model performance across different dimensions of response quality.

Table \ref{tab:baseline_results_zero_shot} displays the zero-shot performance of various models on the OKCV dataset. Bleurt scores for the language-only and open-source Vision-Language Models ranged between 0.47 and 0.59. In contrast, Gemini-1.5-pro achieved a Bleurt score of 0.98. Since Bleurt is trained to assess text similarity in a manner akin to human judgment, this high score suggests that Gemini produces responses that humans perceive as very similar to the reference. This disparity suggests a capability gap between closed-source models like Gemini and the available open-source video-language models.

In contrast, the BartScores were generally low, indicating a low likelihood of a match between the reference and the generated candidate dialogue state. Notably, Gemini scored the lowest BartScore among all models. This suggests that although humans might rate the answers as sufficiently similar, the semantic distance between the generated candidate dialogue state and the ground dialogue state was notably high, with the highest discrepancy observed in responses from the Gemini model. This discrepancy may stem from Bleurt being trained on human ratings, which potentially aligns it more closely with human judgment, whereas BartScore relies solely on BART's pretrained weights and is not fine-tuned to assess semantic similarity in a human-like manner. Despite these differences, both metrics agree that Llama2-7B, when utilizing both captions and external knowledge, achieves the best performance among the open-source models. Conversely, Prometheus-2 scores displayed significant variation, likely due to its sensitivity to the specific rubric used for evaluation. According to Prometheus-2, the Phi-Small model, while utilizing no external knowledge, performed best. Interestingly, even though it was designed to mimic human judgment based on a rubric, Prometheus-2 did not rate the Gemini-pro model highly. Prometheus-2 employs a larger model than Bleurt to mimic human ratings, suggesting that there may be a notable discrepancy between the inferred and ground truth dialogue states.

\section{Conclusion}
This paper introduces OKCV, a dataset that merges the challenges of video processing, dialogue management, and open-domain question answering. As AI systems continue to advance in sophistication, the OKCV dataset serves as a challenging platform to train and test multimodal models in these skills. In this paper, we detail the collection and analysis of the dataset. The dataset comprises 5,986 dialogues with 40,954 dialogue turns across 2,017. Additionally, we established baseline evaluations to gauge the current capabilities of AI models on this dataset.

\section*{Limitations} The OKCV dataset, derived from the videos and transcriptions of the QuerYD dataset, inherits any biases present in its source material. Additionally, biases in the transcriptions can influence the questions generated by GPT-4, subsequently affecting the dialogues produced. While the dataset consists of a variety of dialogues for each video, in particular for longer videos only a small portion of the potential dialogue topics is covered by the dialogue topics within our dataset. Future work could further expand the dialogue topics to an even broader range.
Additionally, the conversation around a certain visually grounded topic normally contains up to four dialogue turns. Particularly, conversations with an in-depth focus on a certain topic with more dialogue turns could be an interesting addition to enlarge the conversation context.

{\small
\bibliography{egbib}

\begin{thebibliography}{50}
\providecommand{\natexlab}[1]{#1}

\bibitem[{kno()}]{knowitxvqa}


\bibitem[{Alamri et~al.(2019)Alamri, Cartillier, Das, Wang, Cherian, Essa, Batra, Marks, Hori, Anderson et~al.}]{alamri2019audio}
Huda Alamri, Vincent Cartillier, Abhishek Das, Jue Wang, Anoop Cherian, Irfan Essa, Dhruv Batra, Tim~K Marks, Chiori Hori, Peter Anderson, et~al. 2019.
\newblock Audio visual scene-aware dialog.
\newblock In \emph{Proceedings of the IEEE/CVF Conference on Computer Vision and Pattern Recognition}, pages 7558--7567.

\bibitem[{Antol et~al.(2015)Antol, Agrawal, Lu, Mitchell, Batra, Zitnick, and Parikh}]{antol2015vqa}
Stanislaw Antol, Aishwarya Agrawal, Jiasen Lu, Margaret Mitchell, Dhruv Batra, C~Lawrence Zitnick, and Devi Parikh. 2015.
\newblock Vqa: Visual question answering.
\newblock In \emph{Proceedings of the IEEE international conference on computer vision}, pages 2425--2433.

\bibitem[{Brown et~al.(2020)Brown, Mann, Ryder, Subbiah, Kaplan, Dhariwal, Neelakantan, Shyam, Sastry, Askell et~al.}]{brown2020language}
Tom Brown, Benjamin Mann, Nick Ryder, Melanie Subbiah, Jared~D Kaplan, Prafulla Dhariwal, Arvind Neelakantan, Pranav Shyam, Girish Sastry, Amanda Askell, et~al. 2020.
\newblock Language models are few-shot learners.
\newblock \emph{Advances in neural information processing systems}, 33:1877--1901.

\bibitem[{Budzianowski et~al.(2018)Budzianowski, Wen, Tseng, Casanueva, Ultes, Ramadan, and Ga{\v{s}}i{\'c}}]{multiwoz}
Pawe{\l} Budzianowski, Tsung-Hsien Wen, Bo-Hsiang Tseng, Inigo Casanueva, Stefan Ultes, Osman Ramadan, and Milica Ga{\v{s}}i{\'c}. 2018.
\newblock Multiwoz--a large-scale multi-domain wizard-of-oz dataset for task-oriented dialogue modelling.
\newblock \emph{arXiv preprint arXiv:1810.00278}.

\bibitem[{Byrne et~al.(2019)Byrne, Krishnamoorthi, Sankar, Neelakantan, Duckworth, Yavuz, Goodrich, Dubey, Cedilnik, and Kim}]{byrne2019taskmaster}
Bill Byrne, Karthik Krishnamoorthi, Chinnadhurai Sankar, Arvind Neelakantan, Daniel Duckworth, Semih Yavuz, Ben Goodrich, Amit Dubey, Andy Cedilnik, and Kyu-Young Kim. 2019.
\newblock Taskmaster-1: Toward a realistic and diverse dialog dataset.
\newblock \emph{arXiv preprint arXiv:1909.05358}.

\bibitem[{Das et~al.(2017)Das, Kottur, Gupta, Singh, Yadav, Moura, Parikh, and Batra}]{das2017visual}
Abhishek Das, Satwik Kottur, Khushi Gupta, Avi Singh, Deshraj Yadav, Jos{\'e}~MF Moura, Devi Parikh, and Dhruv Batra. 2017.
\newblock Visual dialog.
\newblock In \emph{Proceedings of the IEEE conference on computer vision and pattern recognition}, pages 326--335.

\bibitem[{Engin et~al.(2021)Engin, Schnitzler, Duong, and Avrithis}]{engin2021hidden}
Deniz Engin, Fran{\c{c}}ois Schnitzler, Ngoc~QK Duong, and Yannis Avrithis. 2021.
\newblock On the hidden treasure of dialog in video question answering.
\newblock In \emph{Proceedings of the IEEE/CVF International Conference on Computer Vision}, pages 2064--2073.

\bibitem[{Fu et~al.(2021)Fu, Li, Gan, Lin, Wang, Wang, and Liu}]{fu2021violet}
Tsu-Jui Fu, Linjie Li, Zhe Gan, Kevin Lin, William~Yang Wang, Lijuan Wang, and Zicheng Liu. 2021.
\newblock Violet: End-to-end video-language transformers with masked visual-token modeling.
\newblock \emph{arXiv preprint arXiv:2111.12681}.

\bibitem[{Gao et~al.(2022)Gao, Ping, Thattai, Reganti, Wu, and Natarajan}]{gao2022transform}
Feng Gao, Qing Ping, Govind Thattai, Aishwarya Reganti, Ying~Nian Wu, and Prem Natarajan. 2022.
\newblock Transform-retrieve-generate: Natural language-centric outside-knowledge visual question answering.
\newblock In \emph{Proceedings of the IEEE/CVF Conference on Computer Vision and Pattern Recognition}, pages 5067--5077.

\bibitem[{Gao et~al.(2015)Gao, Mao, Zhou, Huang, Wang, and Xu}]{gao2015you}
Haoyuan Gao, Junhua Mao, Jie Zhou, Zhiheng Huang, Lei Wang, and Wei Xu. 2015.
\newblock Are you talking to a machine? dataset and methods for multilingual image question.
\newblock \emph{Advances in neural information processing systems}, 28.

\bibitem[{Garcia et~al.(2020)Garcia, Otani, Chu, and Nakashima}]{garcia2020knowit}
Noa Garcia, Mayu Otani, Chenhui Chu, and Yuta Nakashima. 2020.
\newblock Knowit vqa: Answering knowledge-based questions about videos.
\newblock In \emph{Proceedings of the AAAI conference on artificial intelligence}, volume~34, pages 10826--10834.

\bibitem[{Goyal et~al.(2017)Goyal, Khot, Summers-Stay, Batra, and Parikh}]{goyal2017making}
Yash Goyal, Tejas Khot, Douglas Summers-Stay, Dhruv Batra, and Devi Parikh. 2017.
\newblock Making the v in vqa matter: Elevating the role of image understanding in visual question answering.
\newblock In \emph{Proceedings of the IEEE conference on computer vision and pattern recognition}, pages 6904--6913.

\bibitem[{Gui et~al.(2021)Gui, Wang, Huang, Hauptmann, Bisk, and Gao}]{kat}
Liangke Gui, Borui Wang, Qiuyuan Huang, Alex Hauptmann, Yonatan Bisk, and Jianfeng Gao. 2021.
\newblock Kat: A knowledge augmented transformer for vision-and-language.
\newblock \emph{arXiv preprint arXiv:2112.08614}.

\bibitem[{Guo et~al.(2025)Guo, Yang, Zhang, Song, Zhang, Xu, Zhu, Ma, Wang, Bi et~al.}]{guo2025deepseek}
Daya Guo, Dejian Yang, Haowei Zhang, Junxiao Song, Ruoyu Zhang, Runxin Xu, Qihao Zhu, Shirong Ma, Peiyi Wang, Xiao Bi, et~al. 2025.
\newblock Deepseek-r1: Incentivizing reasoning capability in llms via reinforcement learning.
\newblock \emph{arXiv preprint arXiv:2501.12948}.

\bibitem[{Gupta and Gupta(2022)}]{NEWSKVQA}
Pranay Gupta and Manish Gupta. 2022.
\newblock \href {https://api.semanticscholar.org/CorpusID:246652503} {Newskvqa: Knowledge-aware news video question answering}.
\newblock In \emph{Pacific-Asia Conference on Knowledge Discovery and Data Mining}.

\bibitem[{Honnibal and Montani(2017)}]{spacy2}
Matthew Honnibal and Ines Montani. 2017.
\newblock {spaCy 2}: Natural language understanding with {B}loom embeddings, convolutional neural networks and incremental parsing.
\newblock To appear.

\bibitem[{Hu et~al.(2022)Hu, Shen, Wallis, Allen-Zhu, Li, Wang, Wang, and Chen}]{lora}
Edward~J Hu, Yelong Shen, Phillip Wallis, Zeyuan Allen-Zhu, Yuanzhi Li, Shean Wang, Lu~Wang, and Weizhu Chen. 2022.
\newblock \href {https://openreview.net/forum?id=nZeVKeeFYf9} {Lo{RA}: Low-rank adaptation of large language models}.
\newblock In \emph{International Conference on Learning Representations}.

\bibitem[{Kim et~al.(2024)Kim, Suk, Longpre, Lin, Shin, Welleck, Neubig, Lee, Lee, and Seo}]{prometheus2}
Seungone Kim, Juyoung Suk, Shayne Longpre, Bill~Yuchen Lin, Jamin Shin, Sean Welleck, Graham Neubig, Moontae Lee, Kyungjae Lee, and Minjoon Seo. 2024.
\newblock \href {https://arxiv.org/abs/2405.01535} {Prometheus 2: An open source language model specialized in evaluating other language models}.
\newblock \emph{Preprint}, arXiv:2405.01535.

\bibitem[{Lei et~al.(2018)Lei, Yu, Bansal, and Berg}]{lei2018tvqa}
Jie Lei, Licheng Yu, Mohit Bansal, and Tamara~L Berg. 2018.
\newblock Tvqa: Localized, compositional video question answering.
\newblock In \emph{Empirical Methods in Natural Language Processing}.

\bibitem[{Lei et~al.(2019)Lei, Yu, Berg, and Bansal}]{lei2019tvqa+}
Jie Lei, Licheng Yu, Tamara~L Berg, and Mohit Bansal. 2019.
\newblock Tvqa+: Spatio-temporal grounding for video question answering.
\newblock \emph{arXiv preprint arXiv:1904.11574}.

\bibitem[{Li et~al.(2023)Li, He, Wang, Li, Wang, Luo, Wang, Wang, and Qiao}]{li2023videochat}
KunChang Li, Yinan He, Yi~Wang, Yizhuo Li, Wenhai Wang, Ping Luo, Yali Wang, Limin Wang, and Yu~Qiao. 2023.
\newblock Videochat: Chat-centric video understanding.
\newblock \emph{arXiv preprint arXiv:2305.06355}.

\bibitem[{Lin et~al.(2022)Lin, Xie, Chen, Xu, Zhu, and Yuan}]{lin2022revive}
Yuanze Lin, Yujia Xie, Dongdong Chen, Yichong Xu, Chenguang Zhu, and Lu~Yuan. 2022.
\newblock Revive: Regional visual representation matters in knowledge-based visual question answering.
\newblock \emph{Advances in Neural Information Processing Systems}, 35:10560--10571.

\bibitem[{Lison and Tiedemann(2016)}]{lison2016opensubtitles2016}
Pierre Lison and J{\"o}rg Tiedemann. 2016.
\newblock Opensubtitles2016: Extracting large parallel corpora from movie and tv subtitles.

\bibitem[{Luo et~al.(2022)Luo, Ji, Zhong, Chen, Lei, Duan, and Li}]{clip4clip}
Huaishao Luo, Lei Ji, Ming Zhong, Yang Chen, Wen Lei, Nan Duan, and Tianrui Li. 2022.
\newblock Clip4clip: An empirical study of clip for end to end video clip retrieval and captioning.
\newblock \emph{Neurocomputing}, 508:293--304.

\bibitem[{Mangalam et~al.(2024)Mangalam, Akshulakov, and Malik}]{egoschema}
Karttikeya Mangalam, Raiymbek Akshulakov, and Jitendra Malik. 2024.
\newblock Egoschema: A diagnostic benchmark for very long-form video language understanding.
\newblock \emph{Advances in Neural Information Processing Systems}, 36.

\bibitem[{Marino et~al.(2019)Marino, Rastegari, Farhadi, and Mottaghi}]{marino2019ok}
Kenneth Marino, Mohammad Rastegari, Ali Farhadi, and Roozbeh Mottaghi. 2019.
\newblock Ok-vqa: A visual question answering benchmark requiring external knowledge.
\newblock In \emph{Proceedings of the IEEE/cvf conference on computer vision and pattern recognition}, pages 3195--3204.

\bibitem[{Meng et~al.(2020)Meng, Wang, Han, Sun, Wu, Yan, and Li}]{meng2020openvidial}
Yuxian Meng, Shuhe Wang, Qinghong Han, Xiaofei Sun, Fei Wu, Rui Yan, and Jiwei Li. 2020.
\newblock Openvidial: A large-scale, open-domain dialogue dataset with visual contexts.
\newblock \emph{arXiv preprint arXiv:2012.15015}.

\bibitem[{Nie et~al.(2019)Nie, Wang, Hong, Wang, and Tian}]{nie2019multimodal}
Liqiang Nie, Wenjie Wang, Richang Hong, Meng Wang, and Qi~Tian. 2019.
\newblock Multimodal dialog system: Generating responses via adaptive decoders.
\newblock In \emph{Proceedings of the 27th ACM international conference on multimedia}, pages 1098--1106.

\bibitem[{Oncescu et~al.(2021)Oncescu, Henriques, Liu, Zisserman, and Albanie}]{oncescu2021queryd}
Andreea-Maria Oncescu, Joao~F Henriques, Yang Liu, Andrew Zisserman, and Samuel Albanie. 2021.
\newblock Queryd: A video dataset with high-quality text and audio narrations.
\newblock In \emph{ICASSP 2021-2021 IEEE International Conference on Acoustics, Speech and Signal Processing (ICASSP)}, pages 2265--2269. IEEE.

\bibitem[{OpenAI(2023)}]{gpt4}
OpenAI. 2023.
\newblock \href {https://api.semanticscholar.org/CorpusID:257532815} {Gpt-4 technical report}.

\bibitem[{Pu et~al.(2021)Pu, Chung, Parikh, Gehrmann, and Sellam}]{bleurt2}
Amy Pu, Hyung~Won Chung, Ankur~P Parikh, Sebastian Gehrmann, and Thibault Sellam. 2021.
\newblock Learning compact metrics for mt.
\newblock In \emph{Proceedings of EMNLP}.

\bibitem[{Reichman and Heck(2023)}]{crossmodalretrieval}
Benjamin Reichman and Larry Heck. 2023.
\newblock Cross-modal dense passage retrieval for outside knowledge visual question answering.
\newblock In \emph{Proceedings of the IEEE/CVF International Conference on Computer Vision (ICCV) Workshops}, pages 2837--2842.

\bibitem[{Reichman et~al.(2023)Reichman, Sundar, Richardson, Zubatiy, Chowdhury, Shah, Truxal, Grimes, Shah, Chee et~al.}]{reichman2023outside}
Benjamin~Z Reichman, Anirudh Sundar, Christopher Richardson, Tamara Zubatiy, Prithwijit Chowdhury, Aaryan Shah, Jack Truxal, Micah Grimes, Dristi Shah, Woo~Ju Chee, et~al. 2023.
\newblock Outside knowledge visual question answering version 2.0.
\newblock In \emph{ICASSP 2023-2023 IEEE International Conference on Acoustics, Speech and Signal Processing (ICASSP)}, pages 1--5. IEEE.

\bibitem[{Sanket~Shah and Talukdar(2019)}]{kvqa}
Naganand~Yadati Sanket~Shah, Anand~Mishra and Partha~Pratim Talukdar. 2019.
\newblock Kvqa: Knowledge-aware visual question answering.
\newblock In \emph{AAAI}.

\bibitem[{Schwenk et~al.(2022)Schwenk, Khandelwal, Clark, Marino, and Mottaghi}]{aokvqa}
Dustin Schwenk, Apoorv Khandelwal, Christopher Clark, Kenneth Marino, and Roozbeh Mottaghi. 2022.
\newblock A-okvqa: A benchmark for visual question answering using world knowledge.
\newblock \emph{arXiv}.

\bibitem[{Sellam et~al.(2020)Sellam, Das, and Parikh}]{bleurt1}
Thibault Sellam, Dipanjan Das, and Ankur~P Parikh. 2020.
\newblock Bleurt: Learning robust metrics for text generation.
\newblock In \emph{Proceedings of ACL}.

\bibitem[{Shah et~al.(2019)Shah, Mishra, Yadati, and Talukdar}]{shah2019kvqa}
Sanket Shah, Anand Mishra, Naganand Yadati, and Partha~Pratim Talukdar. 2019.
\newblock Kvqa: Knowledge-aware visual question answering.
\newblock In \emph{Proceedings of the AAAI conference on artificial intelligence}, volume~33, pages 8876--8884.

\bibitem[{Shao et~al.(2023)Shao, Yu, Wang, and Yu}]{shao2023prompting}
Zhenwei Shao, Zhou Yu, Meng Wang, and Jun Yu. 2023.
\newblock Prompting large language models with answer heuristics for knowledge-based visual question answering.
\newblock In \emph{Proceedings of the IEEE/CVF Conference on Computer Vision and Pattern Recognition}, pages 14974--14983.

\bibitem[{Sordoni et~al.(2015)Sordoni, Galley, Auli, Brockett, Ji, Mitchell, Nie, Gao, and Dolan}]{sordoni2015neural}
Alessandro Sordoni, Michel Galley, Michael Auli, Chris Brockett, Yangfeng Ji, Margaret Mitchell, Jian-Yun Nie, Jianfeng Gao, and Bill Dolan. 2015.
\newblock A neural network approach to context-sensitive generation of conversational responses.
\newblock \emph{arXiv preprint arXiv:1506.06714}.

\bibitem[{Speer et~al.(2017)Speer, Chin, and Havasi}]{speer2017conceptnet}
Robyn Speer, Joshua Chin, and Catherine Havasi. 2017.
\newblock Conceptnet 5.5: An open multilingual graph of general knowledge.
\newblock In \emph{Proceedings of the AAAI conference on artificial intelligence}, volume~31.

\bibitem[{Talmor et~al.(2019)Talmor, Herzig, Lourie, and Berant}]{talmor2019commonsenseqa}
Alon Talmor, Jonathan Herzig, Nicholas Lourie, and Jonathan Berant. 2019.
\newblock Commonsenseqa: A question answering challenge targeting commonsense knowledge.
\newblock In \emph{Proceedings of the 2019 Conference of the North American Chapter of the Association for Computational Linguistics: Human Language Technologies, Volume 1 (Long and Short Papers)}, pages 4149--4158.

\bibitem[{Tapaswi et~al.(2016)Tapaswi, Zhu, Stiefelhagen, Torralba, Urtasun, and Fidler}]{tapaswi2016movieqa}
Makarand Tapaswi, Yukun Zhu, Rainer Stiefelhagen, Antonio Torralba, Raquel Urtasun, and Sanja Fidler. 2016.
\newblock Movieqa: Understanding stories in movies through question-answering.
\newblock In \emph{Proceedings of the IEEE conference on computer vision and pattern recognition}, pages 4631--4640.

\bibitem[{Vrande{\v{c}}i{\'c} and Kr{\"o}tzsch(2014)}]{wikidata}
Denny Vrande{\v{c}}i{\'c} and Markus Kr{\"o}tzsch. 2014.
\newblock Wikidata: a free collaborative knowledgebase.
\newblock \emph{Communications of the ACM}, 57(10):78--85.

\bibitem[{Wang et~al.(2017)Wang, Wu, Shen, Dick, and Van Den~Hengel}]{wang2017fvqa}
Peng Wang, Qi~Wu, Chunhua Shen, Anthony Dick, and Anton Van Den~Hengel. 2017.
\newblock Fvqa: Fact-based visual question answering.
\newblock \emph{IEEE transactions on pattern analysis and machine intelligence}, 40(10):2413--2427.

\bibitem[{Wang et~al.(2015)Wang, Wu, Shen, Hengel, and Dick}]{wang2015explicit}
Peng Wang, Qi~Wu, Chunhua Shen, Anton van~den Hengel, and Anthony Dick. 2015.
\newblock Explicit knowledge-based reasoning for visual question answering.
\newblock \emph{arXiv preprint arXiv:1511.02570}.

\bibitem[{Xu et~al.(2023)Xu, Ye, Yan, Shi, Ye, Xu, Li, Bi, Qian, Wang et~al.}]{xu2023mplug}
Haiyang Xu, Qinghao Ye, Ming Yan, Yaya Shi, Jiabo Ye, Yuanhong Xu, Chenliang Li, Bin Bi, Qi~Qian, Wei Wang, et~al. 2023.
\newblock mplug-2: A modularized multi-modal foundation model across text, image and video.
\newblock In \emph{International Conference on Machine Learning}, pages 38728--38748. PMLR.

\bibitem[{Yang et~al.(2022)Yang, Gan, Wang, Hu, Lu, Liu, and Wang}]{yang2022empirical}
Zhengyuan Yang, Zhe Gan, Jianfeng Wang, Xiaowei Hu, Yumao Lu, Zicheng Liu, and Lijuan Wang. 2022.
\newblock An empirical study of gpt-3 for few-shot knowledge-based vqa.
\newblock In \emph{Proceedings of the AAAI Conference on Artificial Intelligence}, volume~36, pages 3081--3089.

\bibitem[{Yuan et~al.(2021)Yuan, Neubig, and Liu}]{bartscore}
Weizhe Yuan, Graham Neubig, and Pengfei Liu. 2021.
\newblock \href {https://proceedings.neurips.cc/paper/2021/file/e4d2b6e6fdeca3e60e0f1a62fee3d9dd-Paper.pdf} {Bartscore: Evaluating generated text as text generation}.
\newblock In \emph{Advances in Neural Information Processing Systems}, volume~34, pages 27263--27277. Curran Associates, Inc.

\bibitem[{Zellers et~al.(2021)Zellers, Lu, Hessel, Yu, Park, Cao, Farhadi, and Choi}]{zellers2021merlot}
Rowan Zellers, Ximing Lu, Jack Hessel, Youngjae Yu, Jae~Sung Park, Jize Cao, Ali Farhadi, and Yejin Choi. 2021.
\newblock Merlot: Multimodal neural script knowledge models.
\newblock \emph{Advances in Neural Information Processing Systems}, 34:23634--23651.

\end{thebibliography}
}

\newpage

\section{Appendix}
\subsection{GPT-4 Prompt for Partial Dialogue Creation}

\begin{figure*}[ht]
\centering
\includegraphics[width=0.5\textwidth]{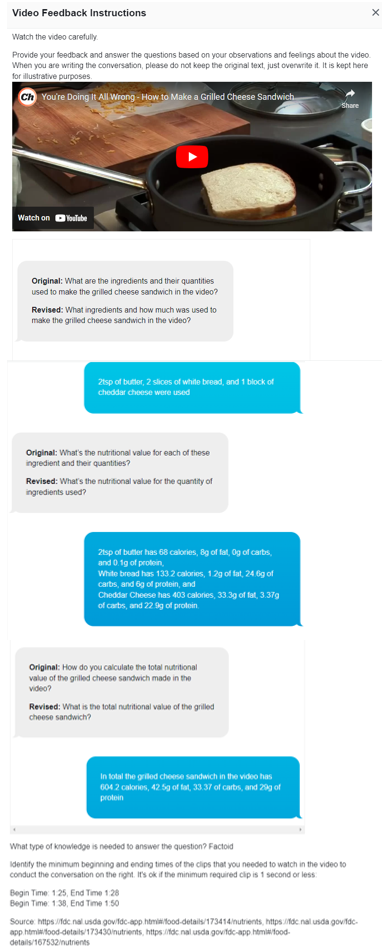}
\caption{Screenshot of example given to Turker of how the collected data should look.}
\label{fig:turk_screenshot3}
\end{figure*}

\begin{figure*}[ht]
\centering
\includegraphics[width=\textwidth]{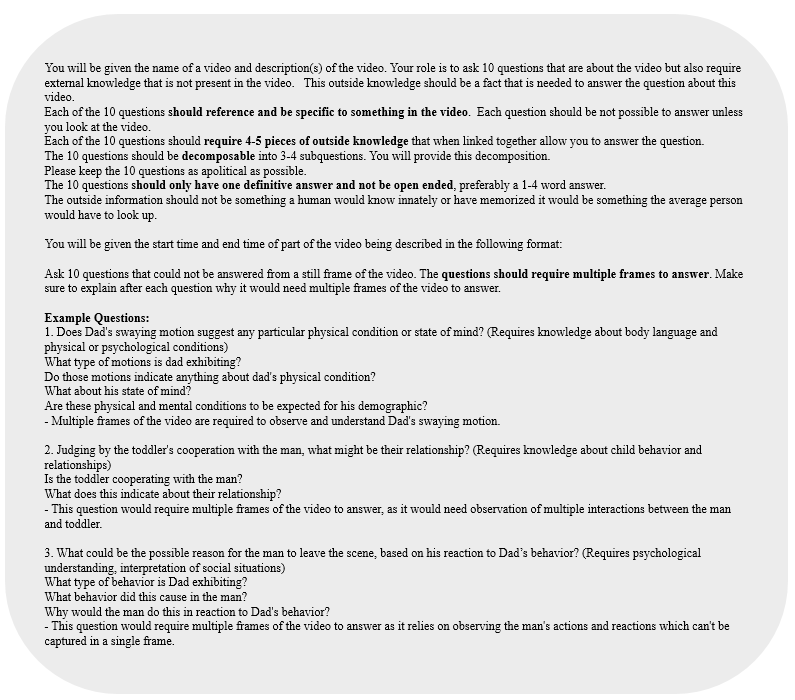}
\caption{System prompt given to GPT-4 to create partial dialogue scripts. Bold is added for emphasis.}
\label{fig:gpt4_prompt}
\end{figure*}

Section 3.2 details how GPT-4 was used to create partial dialogue scripts which were then given to crowdworkers to fill in. Figure \ref{fig:gpt4_prompt} shows the exact system prompt used when GPT-4 was queried. The first part of the prompt are direct instructions for GPT-4. It asks it to create 10 questions for each video. It then asks GPT-4 to make sure that specific parts of the video are referenced in the dialogue so that the video is a necessary part of the task. Then GPT-4 is asked to ask questions that require knowledge outside the video to answer and that the questions be decomposable. It is then asked to decompose the questions. In the middle GPT-4 is asked to ensure that each question should require multiple frames from the video to answer. This would allow for the creation of a dataset that required long video processing. In the last part of the prompt, three example questions are given. Each question is a complex question that is decomposable. Additionally, we provide example justifications for why the questions given are require both outside knowledge (at the end of the question) and multiple frames (at the end of the dialogue).

\subsection{Amazon Mechanical Turk Interface}

\begin{figure*}[t]
\centering
\includegraphics[width=\textwidth]{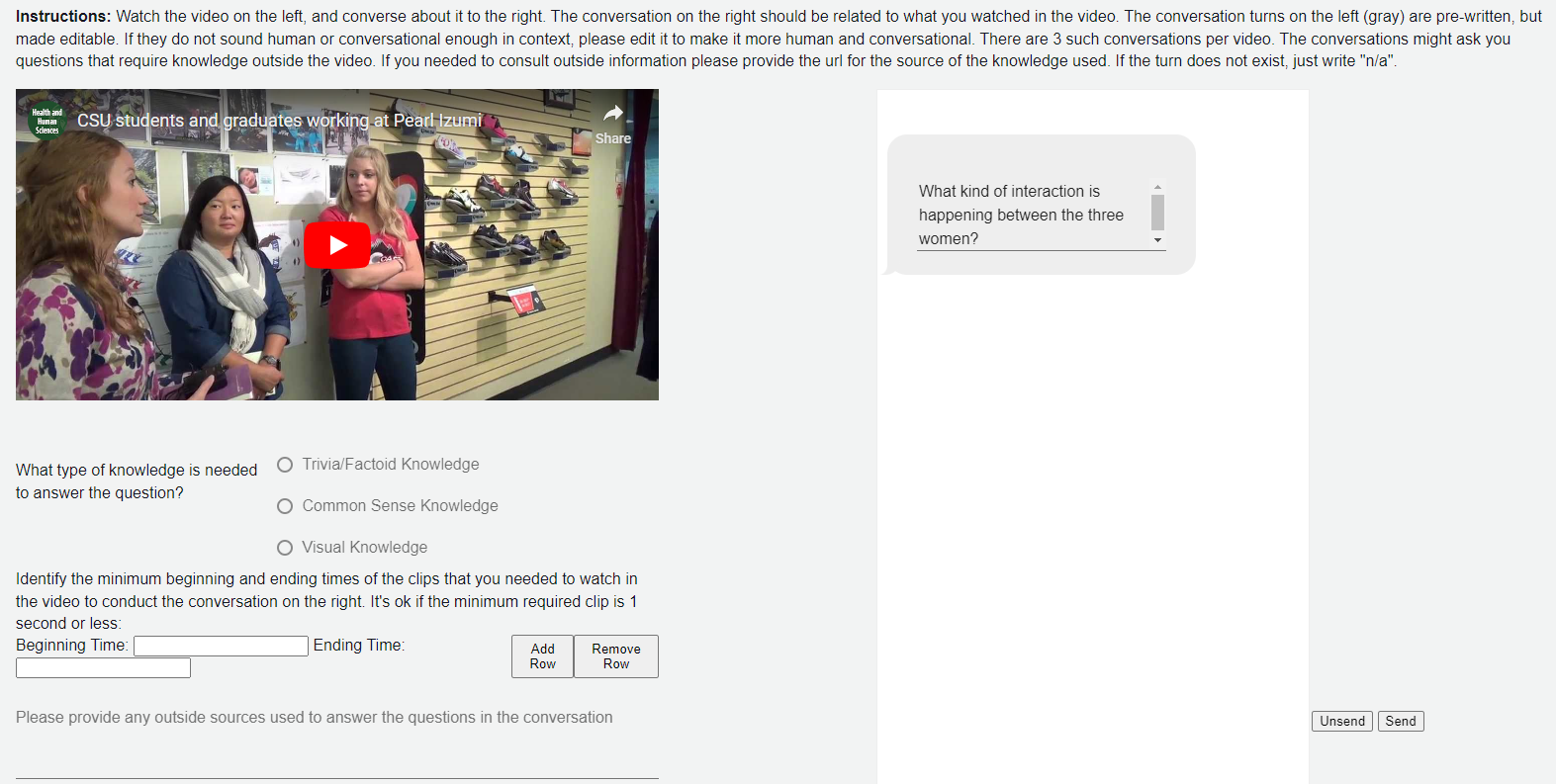}
\caption{Screenshot example of Amazon Mechanical Turk interface at the beginning of the dialogue collection task.}
\label{fig:turk_screenshot1}
\end{figure*}

\begin{figure*}[t]
\centering
\includegraphics[width=\textwidth]{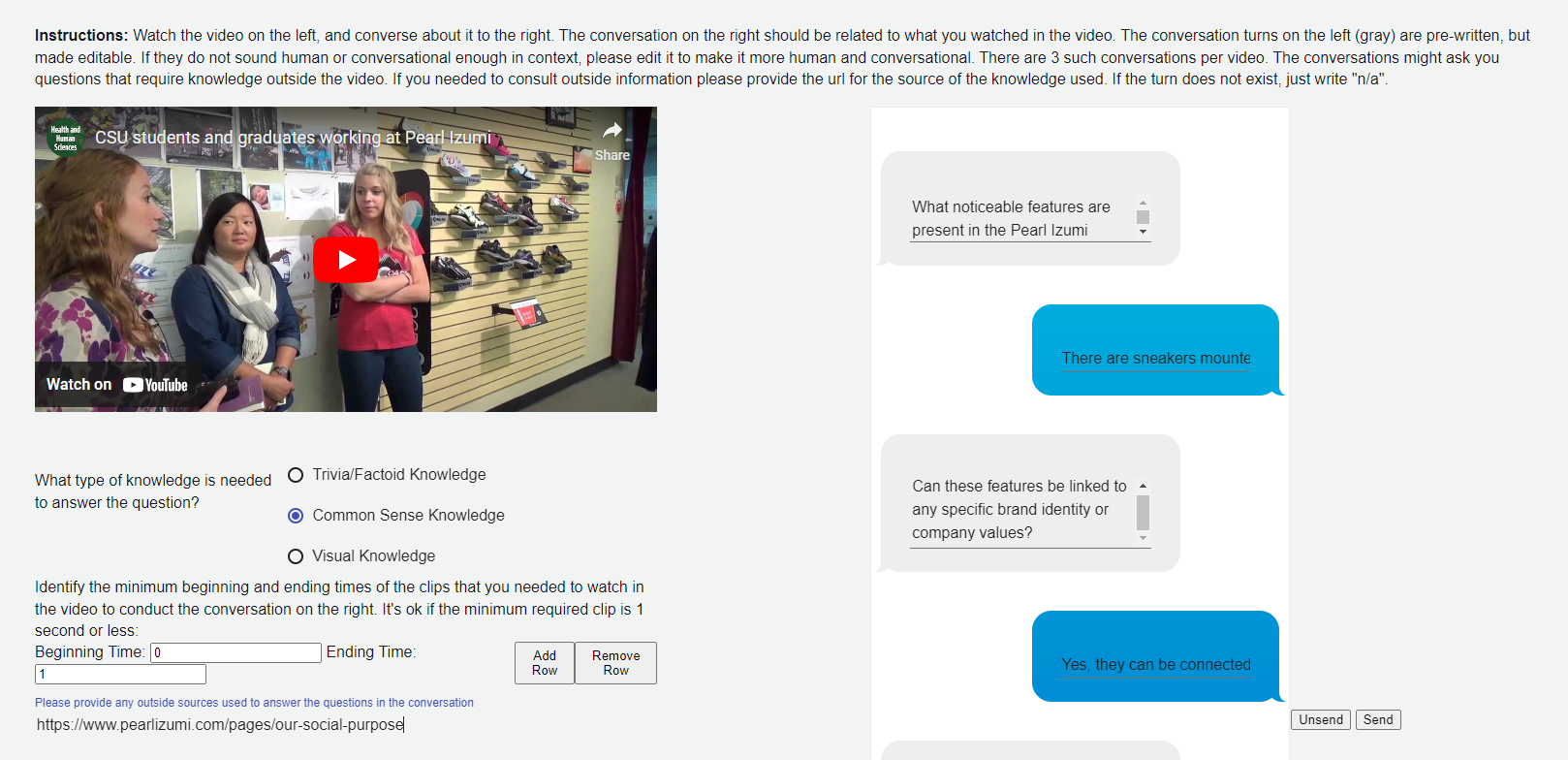}
\caption{Screenshot example of Amazon Mechanical Turker interface at the end of the dialogue collection task.}
\label{fig:turk_screenshot2}
\end{figure*}

Figures \ref{fig:turk_screenshot1} and \ref{fig:turk_screenshot2} visualize the Amazon Mechanical Turk interface built for data collection. On the top written instructions of how to perform the task are displayed. An extended example of how the collected data should look was also given and can be seen in Figure \ref{fig:turk_screenshot3}. On the left-hand column, four different elements of the data collection can be seen. At the top, there is the video that the turker needs to watch in order to complete the task. Below the video, the turker is asked whether the dialogue requires more visual, commonsense, or factoid reasoning. This question is followed below by the temporal certificate element. In this element, the turkers are asked what is the closest begin and end time that they need to complete a dialogue. If there are multiple disjoint times in the video that are needed to answer a question, the turker can click "add time" to add another set of textboxes to fill in. If they accidentally clicked "add row" too many times they can click "remove row". Finally, the last element in this column is a textbox asking the turker to give links to any outside source they used to answer the question in the dialogue.

In the right-hand column, there is a mockup of a texting conversation. Here turkers can fill in the partial dialogues that GPT-4 created. The turker can edit the message they received to be more conversationally fluent. They can send the dialogue chat portion corresponding to the assistant. They can go backwards and forwards in the conversation by clicking "send" and "unsend".

Once the turker fills in the elements on the left and right-hand columns, they are then prompted with the same screen with the same video twice more. Each time though they are prompted with a different dialogue on the right-hand column.

\subsection{Prometheus 2 Rubric Prompt}
\label{app:promethrubric}
\begin{figure*}[t]
\centering
\includegraphics[width=\textwidth]{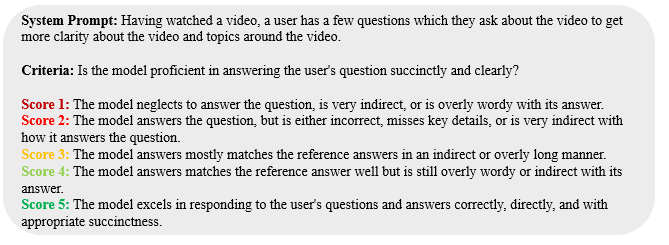}
\caption{Prompt given to Prometheus 2 to evaluate model outputs.}
\label{fig:prometheus_prompt}
\end{figure*}

In Section 5, baseline models are prompted, trained, and evaluated. One of the evaluation methods that we used is Prometheus. This evaluation metric requires a prompted rubric. Figure \ref{fig:prometheus_prompt} shows the rubric that was given to Prometheus to rate dialogue completions. The focus of the rubric was for Prometheus to evaluate how close the generated candidate answer is to a reference answer. However, we wanted it to penalize models for being either indirect or overly lengthy in their responses. A model that answers a question with five paragraphs when the question can be answered in a sentence or two displays a lack of understanding of how dialogues normally flow and should therefore be penalized. On the other hand, if a model is able to answer a question correctly with an appropriate amount of succinctness, it should be rewarded for doing so.

\subsection{Finetuning Results}
\begin{table*}[t!]
\centering
{
\small
\begin{tabular}{c|c|c|c|c}
\toprule
 Backbone Model & Knowledge & Bleurt & BartScore & Prometheus-2  \\
\midrule
LLama2-7B & Nothing & $0.65$ & $0.128$ & $2.28$  \\
\hline
LLama2-7B & Captions & $0.66$ & $0.129$ & $2.19$ \\
\hline
LLama2-7B & Caption+Knowledge & $0.59$ & 0.047 & $2.71$ \\
\hline
LLama3-8B & Nothing & $0.69$ & $0.15$ & $1.97$ \\
\hline
LLama3-8B & Captions & $0.69$ & $0.15$ & $2.03$ \\
\hline
LLama3-8B & Caption+Knowledge & $0.51$ & $0.15$ & $1.28$ \\
\hline 
Mistral-7B-Instruct-v0.3 & Nothing & $0.60$ & $0.035$ & $2.53$  \\
\hline
Mistral-7B-Instruct-v0.3 & Captions & 0.60 & $0.036$ & $2.55$  \\
\hline
Mistral-7B-Instruct-v0.3 & Captions+Knowledge & $0.60$ & $0.036$ & $2.62$  \\
\bottomrule
\end{tabular}
}
\caption{Fine-tuning results of baselines on the OKCV dataset.}
\label{tab:baseline_results_unfiltered}
\end{table*}

All model fine-tuning was performed on a server with eight NVIDIA A40 GPUs with most experiments taking up to a few hours to perform. The models were fine-tuned using LORA with a $32$-bit paged ADAMW optimizer and a learning rate of $2e-5$ \cite{lora}. The data was randomly split with $80\%$ of the collected dialogues forming the train split and $20\%$ forming the test split. 
We evaluated the backbone models under various knowledge configurations, starting with a "no knowledge" setting. In this scenario, the model was provided only with the current dialogue state and tasked with generating the subsequent state, both in zero-shot and fine-tuned configurations. This experimental setup allowed us to characterize the utility of the world knowledge of these models in dialogue completion.

Current baseline models exhibit varied results when fine-tuned on the OKCV dataset, as reflected in Table \ref{tab:baseline_results_unfiltered}. After fine-tuning, Bleurt scores for all models increase compared to their zero-shot performance. Similarly, BartScore improves for all models except the Mistral models, where a slight decrease is seen. However, Prometheus-2 scores decrease across all models following fine-tuning when compared to the results from zero-shot prompting. Both Bleurt and Prometheus-2 are trained on human judgment data. However, they apply the data differently. Bleurt, is specifically trained to mimic human judgment for text matching. Prometheus-2, on the other hand, is trained to evaluate responses based on a user-defined rubric, making its assessments more rubric-sensitive. The general increase in Bleurt and Bart scores suggests that fine-tuning likely enhances model performance on the OKCV dataset, despite the mixed results from the Prometheus 2 metric.

The different forms of knowledge applied during the fine-tuning of models on the OKCV dataset yield mixed results. Bleurt scores indicate that models lacking additional knowledge perform comparably to those with access to video captions. However, models that incorporate both captions and external knowledge retrieved from WikiData tend to underperform compared to those accessing only captions or solely the dialogue prompts. The BartScores similarly display this trend in the Llama2 models, while showing that added knowledge does little to improve performance in other fine-tuned models. Conversely, Prometheus-2 scores show a $15.8\%$ improvement for Llama2 when both captions and external Wikidata knowledge is provided, a $35\%$ decrease for Llama3 when both are used, and a modest $3.6\%$ increase for Mistral under the same conditions. These varied outcomes suggest that the captions are likely useful knowledge augmentations for these language models in completing the dialogues whereas the knowledge from WikiData may be neutral at best. Future work should explore knowledge retrieval methods more tailored to video and different knowledge bases.

\end{document}